%% file: main.tex
\documentclass[letterpaper]{article} 
\usepackage{formatting}  
\usepackage{times}  
\usepackage{helvet}  
\usepackage{courier}  
\usepackage{url}  
\usepackage{graphicx}  
\usepackage{caption}
\usepackage{amsmath}
\usepackage{graphicx}
\usepackage{subfig}
\usepackage{multirow}
\usepackage[title]{appendix}

\begin{document}

\title{Interactive Learning of Environment Dynamics for Sequential Tasks}
\author{
Robert Loftin\thanks{Work done in part while authors were with Borealis AI.} \\
North Carolina State University \\
Raleigh, NC 27695 \\
\texttt{rtloftin@ncsu.edu}
\And
Bei Peng\footnotemark[1]\\
Washington State University \\
Pullman, WA 99164, USA \\
\texttt{bei.peng@wsu.edu}
\And
Matthew E. Taylor \\
Borealis AI \\
Edmonton, AB T5J 3G2, Canada \\
\texttt{matthew.taylor@borealisai.com}
\AND
Michael L. Littman \\
Brown University \\
Providence, RI 02912, USA \\
\texttt{mlittman@cs.brown.edu}
\And
David L. Roberts \\
North Carolina State University \\
Raleigh, NC 27695, USA \\
\texttt{robertsd@csc.ncsu.edu}
}
\maketitle

\begin{abstract}
	
    In order for robots and other artificial agents to efficiently learn to perform useful tasks defined by an end user, they must understand not only the goals of those tasks, but also the structure and dynamics of that user's environment.  While existing work has looked at how the goals of a task can be inferred from a human teacher, the agent is often left to learn about the environment on its own.  To address this limitation, we develop an algorithm, \emph{Behavior Aware Modeling} (BAM), which incorporates a teacher's knowledge into a model of the transition dynamics of an agent's environment.  We evaluate BAM both in simulation and with real human teachers, learning from a combination of task demonstrations and evaluative feedback, and show that it can outperform approaches which do not explicitly consider this source of dynamics knowledge.
    
\end{abstract}

\section{Introduction}
\label{sec:intro}
    
    A key goal of interactive learning research is to allow robots and other artificial agents to leverage human knowledge to more efficiently learn useful tasks.  Such agents often require large amounts of data to learn tasks on their own, data which, while easy to generate in simulation, can be expensive to collect in real-world settings.
    Through interactive learning, an agent can rely on human knowledge in addition to its own, limited experience.  The challenge is learning from modes of communication that are natural for users (whom we will refer to as the \emph{teachers}) who lack expertise in programming or AI.  Here we consider two such modes: task demonstrations~\cite{argall2009survey} and evaluative feedback~\cite{knox2013tamer}.  As these types of data are expensive to collect, we need to extract as much information as possible from interaction with a teacher.  In this work, we develop an algorithm, \emph{Behavior Aware Modeling} (BAM), which incorporates data provided by a teacher for several different tasks into a single model of the agent's environment, which allows the teacher's knowledge to be shared across all of the tasks the agent is learning.
    
    A teacher's understanding of their environment is implicit in the way they choose to perform different tasks, and BAM allows an agent to build this knowledge into its own model of the world.  There are many settings where an agent must learn to operate in an environment that its human teachers would already know well.  For example, a robot that transports materials around a hospital needs to learn the layout of the building, and how the people within it will behave (e.g. will someone pushing a gurney make way for the robot?).   A smart home system will be more effective if it knows how the HVAC system affects different parts of the house, and where the occupants are likely to be at a given time.  While an agent could acquire such information on its own, this might be expensive or even dangerous (e.g. the robot blocking a gurney to see what happens).  This work shows how such knowledge can be acquired from human teachers, in combination with an agent's own experience. 
    
    BAM treats the problem of learning multiple tasks as a set of Markov decision processes, each with the same transition dynamics, and captures the teacher's understanding of the environment with a \emph{model}~\cite{brafman2002rmax,deisenroth2011pilco} of these dynamics, while learning separate cost functions encoding the goals of each of the tasks. BAM can be seen as a generalization of \emph{Inverse Reinforcement Learning} (IRL), which learns to perform tasks by identifying the cost functions defining those tasks~\cite{abbeel2004apprenticeship,vroman2014mlirl}.  The key difference between BAM and IRL is that IRL relies solely on the agent's own understanding of the transition dynamics, while BAM also exploits the teacher's knowledge of these dynamics.  BAM combines its own experience and prior knowledge of the dynamics with that provided by the teacher, and because its dynamics model is shared across tasks, the policy it learns for one task can incorporate data provided for another. 
    
    In addition to deriving a novel algorithm for learning tasks and dynamics from demonstrations and evaluative feedback, we also present empirical results comparing BAM against existing approaches to learning from such data.  We evaluate BAM both with simulated teachers and with human teachers through a large-scale, web-based user study.  Our results demonstrate that by explicitly capturing a teacher's understanding of the environment, BAM can significantly reduce the total effort required to teach an agent to perform a collection of tasks.  Interactive learning is limited by the amount of information a human teacher can provide, and so by reducing the effort required on the part of the teacher BAM may allow for interactive learning to be applied to more complex and useful tasks than would otherwise be feasible.

\section{Related Work}
\label{sec:related}
    
    This work considers the case where a human teacher demonstrates a set of tasks, and then provides evaluative feedback while the agent attempts to perform these tasks itself.  As there is no observable cost function, standard reinforcement learning methods cannot be directly applied.  For learning from demonstrations, the simplest approach would be \emph{behavioral cloning}~\cite{Bain1995cloning,pomerleau1989alvinn}, where a supervised learning algorithm is used to find a mapping from states to actions based of the teacher's actions.  Behavioral cloning, however, often struggles to find robust policies for sequential tasks~\cite{ross2011dagger,atkeson1997robots}.  By using knowledge of the dynamics of the environment, and finding cost functions describing the tasks being taught, inverse reinforcement learning can produce policies which are robust over longer time horizons, and which generalize to states not encountered during training~\cite{abbeel2004apprenticeship,ng2000inverse}.
    
    The dynamics knowledge on which IRL depends is sometimes provided in the form of explicit state transition probabilities~\cite{ramachandran2007bayesian,syed2008lpal}, but in more realistic settings the agent must acquire this information through its own interaction with the environment~\cite{abbeel2004apprenticeship,bloem2014infinite,boularias2011relative}.  In contrast to existing IRL approaches, the algorithm described here learns about the dynamics based on the teacher's demonstrations and feedback.  BAM interprets teacher actions in much the same way as many existing IRL algorithms~\cite{ramachandran2007bayesian,neu2007gradient,vroman2014mlirl}, assuming that each action is sampled from a Boltzmann distribution based on the expected return values, or $Q$-values, for possible actions in the current state.  BAM is most closely related to Maximum-Likelihood IRL (ML-IRL)~\cite{vroman2014mlirl}, in which these $Q$-values are assumed to be computed through a differentiable, \emph{soft} value iteration process, and costs are found via gradient ascent on the log-likelihood of the data.
    
    We note that other recent work has taken a similar approach to ours, learning both cost functions and dynamics parameters from human demonstrations via a maximum-likelihood IRL algorithm~\cite{herman2016serd}.  In contrast to that work however, this work demonstrates that incorporating teacher knowledge into an agent's dynamics model is beneficial in the context of real-time, interactive learning.  We also consider the problem of learning multiple tasks (with multiple cost functions) simultaneously.  Finally, our algorithm can incorporate positive and negative feedback from the teacher, relying on a variation of the Bayesian interpretation of feedback developed in~\cite{loftin2016sabl}.

\section{Algorithms}
\label{sec:algorithms}

    The behavior-aware modeling algorithm assumes that information coming from the teacher, both demonstrations of the target tasks and evaluative feedback given in response to the agent's behavior, depends on some state-action value function $Q$ known only to the teacher.  This $Q$-function itself depends on the cost function $C$ which the teacher associates with the current task, as well as the teacher's internal model $T$ of the dynamics of the environment (which we assume is equivalent to the true dynamics).  We then assume that the $Q$-function for a task is computed via a finite number of steps of the same \emph{soft} value iteration used in ML-IRL, defined by
    \begin{align}
	\label{eqn:soft_bellman}
	Q_{t}(s, a) &= -C(s) + \sum_{s^\prime \in S} T(s, a, s^\prime) V_{t-1}(s^\prime) \\
	V_{t}(s) &= \sum_{a \in A} Q_{t}(s, a)\frac{1}{Z_{t}(s)}e^{\beta Q_{t}(s, a)},
	\end{align}
	where $Z(s) = \sum_{a \in A} e^{\beta Q_{t}(s, a)}$ is a normalization term.  This soft value iteration has the advantage of being differentiable, and accounts for the teacher's potentially suboptimal behavior. It also allows the teacher's actions and feedback to depend on states that it would not under optimal planning.  
	
	The BAM algorithm estimates $T$ and the cost $C_i$ for each task $i$ by maximizing their probability given the data provided by the teacher (feedback and demonstrated actions), as well as the agent's direct observations of state transitions.  We divide the training data provided to BAM into a set $D_T$ containing the teacher's actions and feedback, and a set $D_E$ containing the state transitions observed when either the teacher or the agent takes an action.  We further divide $D_T$ into sets $D_{T}^{i}$ for each task $i$ being taught.  $D_E$ consists of state transitions $\sigma = \lbrace s, a, s' \rbrace$, while each set $D_{T}^{i}$ consists of state-action pairs $\varsigma = \lbrace s, a \rbrace$ and feedback events $\vartheta = \lbrace s, a, f \rbrace$.  Similar to Bayesian IRL~\cite{ramachandran2007bayesian}, BAM assumes that a teacher samples actions from a Boltzmann distribution such that $p(a \vert s) \propto e^{\beta Q(s, a)}$.

    To incorporate positive and negative feedback, we employ a version of the SABL feedback model developed in~\cite{loftin2016sabl}.  This version, which we refer to as \emph{Advantage}-SABL (A-SABL), defines the probability of receiving a positive or negative feedback signal from the teacher in terms of the \emph{advantage} of the most recent action.  Specifically, we define the advantage of action $a_t$ as $\delta_t = Q(s_t, a_t) - \frac{1}{\vert A\vert}\sum_{a' \in A} Q(s_t, a')$ (the advantage under a random policy).  The probabilities of receiving positive feedback $f^+$ or negative feedback $f^-$ for $a_t$ are then
    \begin{align}
    p(f_t \!=\! f^+ \vert s_t, a_t) &= (1\! - \!\mu^+)\left[(1 - 2\epsilon)\sigma(\alpha\delta_t) + \epsilon\right] \\
    p(f_t \!=\! f^- \vert s_t, a_t) &= (1\! - \!\mu^-)\left[(1 - 2\epsilon) \sigma(-\alpha\delta_t) + \epsilon\right],
    \end{align}
    where $\sigma(x) = 1 / (1 + e^{-x})$.  The probability of receiving no feedback is simply $1 - p(f_t = f^+ \vert s_t, a_t) - p(f_t = f^- \vert s_t, a_t)$.  $\mu^+$ and $\mu^-$ are tunable parameters that define the probability of receiving \emph{explicit} feedback given that the teacher interprets an action as correct or incorrect, while $\epsilon$ is the teacher's error rate, and $\alpha$ is a scale factor.  

    \subsection{Behavior Aware Modeling}
	\label{sec:bam-algorithm}

	BAM works with a parametric space of dynamics models and cost functions, and computes maximum likelihood estimates of the parameters $\theta$ of the dynamics model $T_{\theta}$, as well as the parameters $\phi_i$ of the cost functions $C_{\phi_i}$ for each of the tasks being taught.  Both the dynamics parameters $\theta$ and the parameters of $\phi_i$ of each cost function (which we write compactly as $\hat{\phi}$) are learned via gradient ascent on their log-probability, that is, BAM maximizes the objective function:
	\begin{multline}
	\label{eqn:bam}
	L(\theta, \hat{\phi} ; D_T, D_E) = \sum_{i = 1}^{\vert D_T \vert} \left[\ln p(D_{T}^{i} \vert Q_{\theta, \phi_i})  + \ln n(\phi_i)\right] \\ + \sum_{\sigma \in D_E} \ln T_{\theta}(s, a, s') + \ln m(\theta),
	\end{multline}
	where $Q_{\theta, \phi_i}$ is the $Q$-function computed under the model $T_{\theta}$, for the task defined by $C_{\phi_i}$, and where $n(\phi)$ and $m(\theta)$ are regularization terms.  The most computationally difficult part of the optimization is the gradient of $\ln p(D_{T}^{i} \vert Q_{\theta, \phi_i})$ w.r.t. $\theta$ and $\hat{\phi}$.  The gradient w.r.t. $\theta$,
	\begin{multline}
	\nabla_{\theta} \ln p(D_{T}^{i} \vert Q_{\theta, \phi_i}) = 
	\\ \sum_{\varsigma \in D_{T}} \!\!\left[\beta \nabla_{\theta}Q_{\theta, \phi_i}(s, a) - \sum_{a' \in A} \pi(s, a') \beta \nabla_{\theta} Q_{\theta, \phi_i}(s, a') \right] 
	\\  + \!\!\sum_{\vartheta \in D_{T}} \!\!\nabla_{Q_{\theta, \phi_i}(s)}\left[ \ln p(f \vert s, a, Q_{\theta,\phi_i}(s)) \right] \frac{\partial}{\partial \theta} Q_{\theta, \phi_i}(s),
	\end{multline}
	where $\pi(s, a) = \frac{1}{Z(s)}e^{Q_{\theta,\phi_i}(s, a)}$, depends on $\nabla_{\theta}Q_{\theta,\phi_i}(s, a)$ and on $\frac{\partial}{\partial \theta} Q_{\theta,\phi_i}(s)$, the Jacobian of the $Q$-values for state $s$ w.r.t. $\theta$.  The gradient for $\phi_i$ takes the same form.  We assume that $Q_{\theta, \phi_i} = Q^{\tau}$, where $Q^{\tau}$ is the $\tau$th step of a soft value iteration process.  For state $s$ and action $a$ we then have 
	\begin{align}
	&\begin{aligned}
	\nabla_{\theta} Q^{t} (s, a) &= -C_{\phi_i}(s) \\
	+ \sum_{s' \in S} T_{\theta}(s, & a, s') \left[ \nabla_{\theta} \ln T_{\theta}(s, a, s') + \nabla_{\theta} V^{t-1}(s')\right], 
	\end{aligned}\\
	&\begin{aligned}
	\nabla_{\phi_i} Q^{t} (s, a) = & -\nabla_{\phi_i} C_{\phi_i}(s) \\
	& + \sum_{s' \in S} T_{\theta}(s, a, s') \nabla_{\phi_i} V^{t-1}(s'),
	\end{aligned}
	\end{align}
	and for both $\theta$ and $\phi_i$ we have
	\begin{multline}
	\nabla V^{t}(s) = \\ \sum_{a \in A} \pi^{t} (s, a) \left[(1 - \beta Q^{t}(s, a) + \beta V^{t} (s)) \nabla Q^{t}(s,a)\right],
	\end{multline}
	where $\pi^{t}(s, a) \propto e^{-\beta Q^{t}(s, a)}$, that is, $\pi^{t}$ is the Boltzmann action distribution for state $s$.  In our implementation, we first accumulate the likelihood terms for all demonstrated actions and feedback before computing this gradient.  Rather than maximizing Equation~\ref{eqn:bam} for the $\theta$ and $\hat{\phi}$ simultaneously, we have found empirically that alternating between optimizing $\hat{\phi}$ and optimizing $\theta$ is more efficient and reliable in finding good estimates of the dynamics and cost-functions.

\section{Simulated Teacher Experiments}
\label{sec:simulation}
    
    \begin{figure*}[!t]
    	\centering
    	\subfloat[{\scriptsize Navigation - Two Rooms}]{
    		\label{fig:two_rooms}
    		\includegraphics[width=0.5\columnwidth]{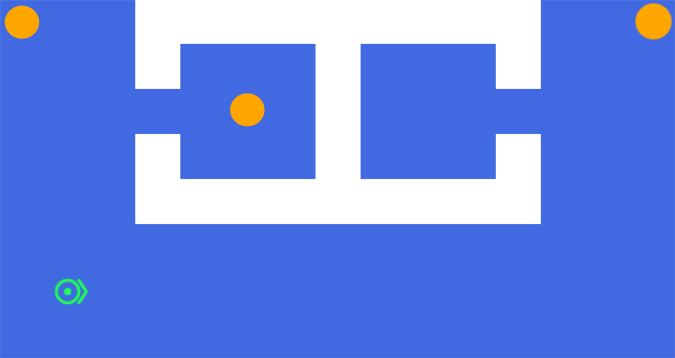}
    	}
    	\subfloat[{\scriptsize Navigation - Doorway}]{
    		\label{fig:doors}
    		\includegraphics[width=0.4\columnwidth]{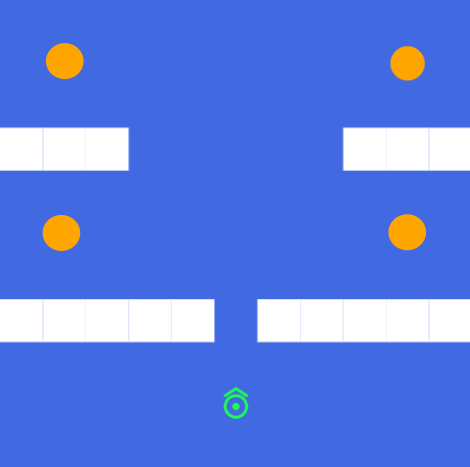}
    	}
    	\subfloat[{\scriptsize Farming - Two Fields}]{
    		\label{fig:two_fields}
    		\includegraphics[width=0.5\columnwidth]{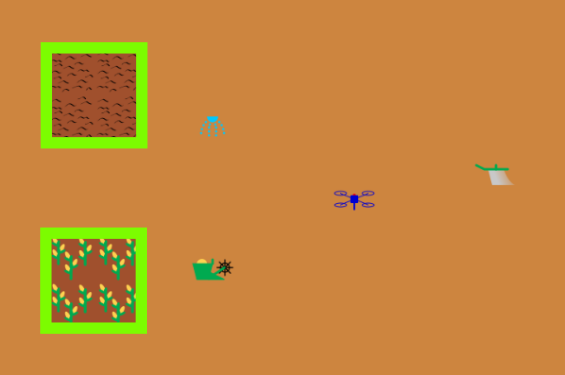}
    	}
    	\subfloat[{\scriptsize Farming - Three Fields}]{
    		\label{fig:three_fields}
    		\includegraphics[width=0.5\columnwidth]{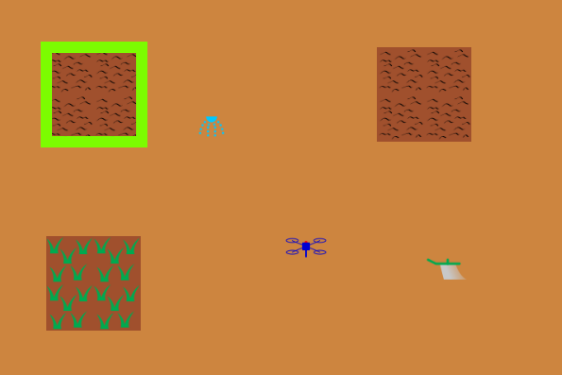}
    	}
    	\caption{{\small Four of the learning environments used in both the simulated teacher and human subjects experiments.  The goal locations are highlighted with either orange circles or green squares.}}
    	\label{fig:environments_human}
    \end{figure*}
    
    To understand how learning dynamics can reduce the effort needed to teach a set of behaviors, we compare BAM against two other approaches to interactive learning, using data generated by a simulated teacher.  We consider three classes of learning problems, which we refer to as \emph{domains}, with discrete state and action spaces.  For each domain, we define multiple \emph{environments}, where each environment is defined by its specific transition probabilities, which are initially unknown to the learning agents.  Within each environment, we define one or more \emph{tasks}, each defined by different cost function, which are also unknown.  In our experiments, an individual agent attempts to learn all the tasks within a single environment.  Each environment also defines a space of possible dynamics models and cost functions which the agents must choose from.  In all three domains, the true cost functions were zero everywhere except for the goal states.
    
    The first domain, which we refer to as \emph{navigation} (see Figures~\ref{fig:two_rooms}, \ref{fig:doors}), is a grid world in which any grid cell may be blocked by an obstacle.  Each task in this domain is defined by a goal location, while the dynamics are defined by the set of cells that are blocked by obstacles.  The space of cost functions has one parameter for each cell in the grid, potentially allowing obstacles to be represented as high-cost cell.  The dynamics model also has one parameter per cell, and a transition into cell $i$ fails with probability $1 / (1 + e^{-\theta_i})$.
    
    The two other domains are grid worlds, in which each grid cell has an additional feature which affects the transition dynamics, such that multiple states correspond to single cell.  In the \emph{farming} domain (see Figures~\ref{fig:two_fields}, \ref{fig:three_fields}), the agent may carry one of three farm implements (a plow, a sprinkler, or a harvester), and the implement it carries determines which cells it is able to enter.  Cells representing dirt fields require the plow, while cells with immature crops require the sprinkler, and cells with fully grown crops require the harvester.  There are also cells in which the agent can pick up each implement. Tasks are defined by groups of cells that the agent must reach.  The space of dynamics models is a space of mappings from each implement to the probability that it works on a certain type of cell, that is, a mapping from the three implements to distributions over the three cell types.  
    
    In the \emph{gravity} domain there are four possible gravity directions, and the agent cannot move in the direction opposite to the current gravity.  A task is defined by a goal cell, and certain cells allow the agent to change the gravity direction so that it can reach the goal.  Each of these cells has a color which determines how it changes the direction of the gravity, and the space of dynamics models is a space of mappings from colors to distributions over the gravity directions.  The cost function spaces for the farming and gravity domains allow independent cost for every grid cell (but not every state).

    \subsection{Alternative Algorithms}
    \label{sec:alternatives}
   
    We wish to determine whether incorporating teacher knowledge into the agent's dynamics model allows for more efficient learning than using a model built solely from directly observed state transitions.  As the BAM algorithm can be viewed as a generalization of maximum-likelihood IRL to the problem of learning transition dynamics, we compare BAM against a version of ML-IRL that uses a dynamics model which does not incorporate information from the teacher.  This \emph{model-based} IRL algorithm first finds a maximum-likelihood model of the transition dynamics based only on the observed transitions in $D_E$, and then uses ML-IRL to find the cost function for each task.  Model-based IRL selects its dynamics model from the same spaces of models as BAM does, and selects its actions greedily.

    Both BAM and ML-IRL generalize to states for which they have received no teacher data by learning cost functions that describe the tasks being taught.  As this approach may be ineffective, or even counterproductive in some cases, we also compare BAM against a \emph{behavioral cloning} algorithm, one that uses a tabular representation that does not generalize between different states.  To incorporate both feedback and demonstrations, our behavioral cloning algorithm finds a table of $Q$-values for each state-action pair, rather than a policy, and selects its actions greedily (as with BAM and model-based IRL).  These $Q$-values, however, can be interpreted simply as the log-probabilities of each action, with the greedy policy selecting the most probable action.

    \subsection{Learning from Demonstration}
    \label{sec:sim_results}
    
    \begin{figure*}[!htb]
    	\centering
    	\includegraphics[width=2\columnwidth]{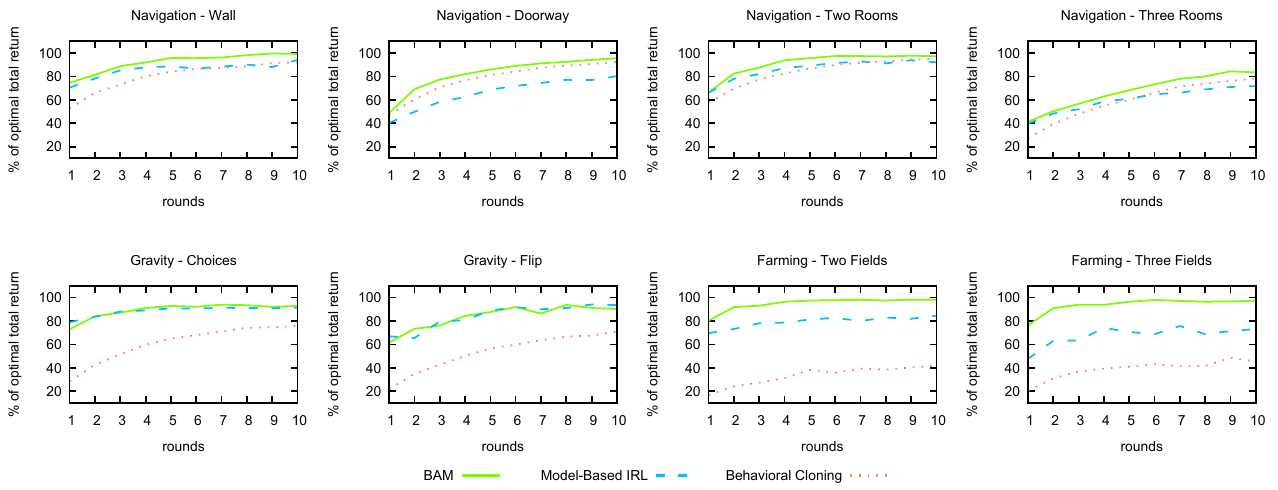}
    	\caption{{\small The total return of the policies learned by BAM, model-based IRL, and behavioral cloning, as a percentage of the total return for the optimal policies.  Curves are averages over 50 separate agents learning from scratch.}}
    	\label{fig:performance_goal}
    \end{figure*}
    
    \begin{figure*}[!htb]
	    \centering
	    \includegraphics[width=2\columnwidth]{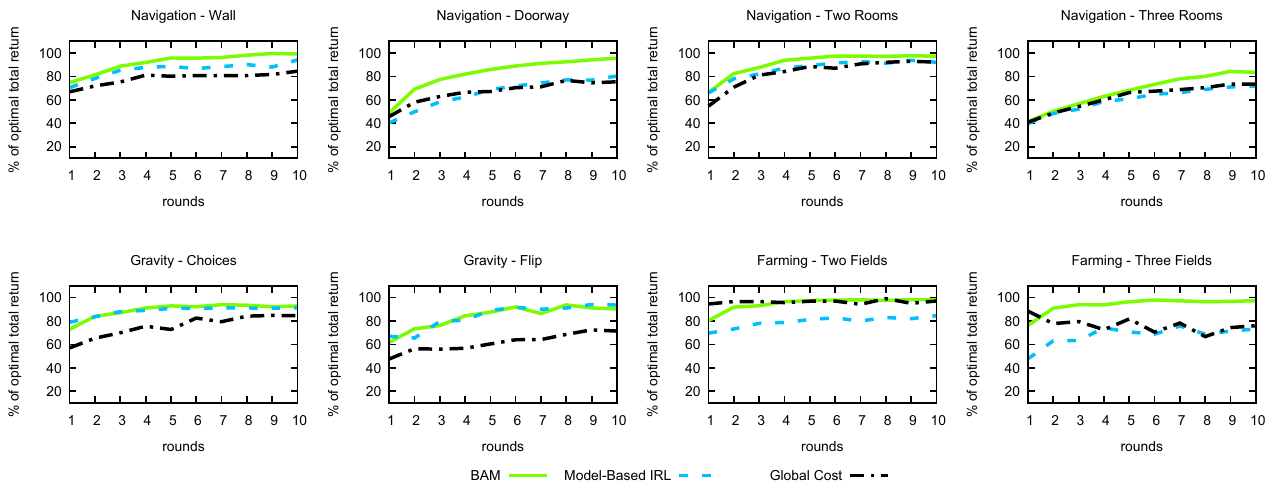}
	    \caption{{\small The total return of the policies learned by BAM, model-based IRL, and model-based IRL with global costs, as a percentage of the total return for the optimal policies.  Curves are averages over 50 separate agents learning from scratch.}}
	    \label{fig:performance_common}
    \end{figure*}

    In our first set of experiments, the simulated teacher only provided demonstrations, and the agent did no exploration on its own.  In each \emph{round}, a single demonstration of the optimal policy was given for each task in the current environment, after which the agent updated its policies for each of these tasks to incorporate the new demonstrations.  To best reflect the behavior of real teachers, each demonstration was terminated after the goal state was reached, with a final action allowing the agent to observe the teacher remaining at the goal.  We evaluated the set of policies an agent learned at a given point in time by their \emph{total return}, that is, the sum of the expected returns of policies for each task.  The expected return of was estimated by running 50 simulated episodes following a policy (the agent did not see these episodes).
    
    Figure~\ref{fig:performance_goal} shows the total return of the policies learned by each of the algorithms, as a percentage of the total return of the optimal policies, and plotted against the number of rounds of demonstrations.  We can see that BAM substantially outperforms both algorithms in the farming environments, while outperforming behavioral cloning in the gravity environments, and model-based IRL in the Doorway environment.  Furthermore, we can see that BAM dominates the other algorithms in that it always performs at least as well as the strongest alternative, while model-based IRL and behavioral cloning perform inconsistently.
    
    The apparent advantage of the BAM algorithm over model-based IRL in the Doorway environment (see Figure~\ref{fig:doors}) is of particular interest.  The initial state in this environment is randomly chosen from the bottom three rows of the grid, such that the agent must go through the doorway to reach any of the goals.  While the walls are never encountered during the optimal demonstrations, BAM is able to identify the location of the doorway and share this information across all four tasks.  This example suggests that inferring a teacher's dynamics model may be an effective alternative to extracting intermediate policies, commonly represented as \emph{options}~\cite{sutton1999smpd} in RL, from human demonstrations.  Rather than learning policy for going through the doorway, BAM can find a dynamics model which leads the agent to use the doorway, thus capturing the same behavior.   Model-based IRL, however, must encode the doorway within its task-specific cost functions, and so must learn this behavior separately for each task. 
    
    The BAM algorithm also has an advantage in the farming environments.  In both of these environments, the agent must move away from the target field to retrieve the plow.  While the agent never directly observes the fact that the other implements do not work for the target field, it can infer this outcome from the fact that the teacher went out of their way to reach the correct machine.  Finally, we attribute the relatively strong performance of behavioral cloning in the navigation domains to the fact that the cloning agent follows a random policy until it reaches a state it has observed before (which does not take very long), after which it knows the optimal policy all the way to the goal.  Even so, BAM still performs at least as well as behavioral cloning in these environments.
    
    \subsection{Global Cost Functions}
    \label{sec:global_cost}
    
    \begin{figure*}[!htb]
        \centering
    	\includegraphics[width=2\columnwidth]{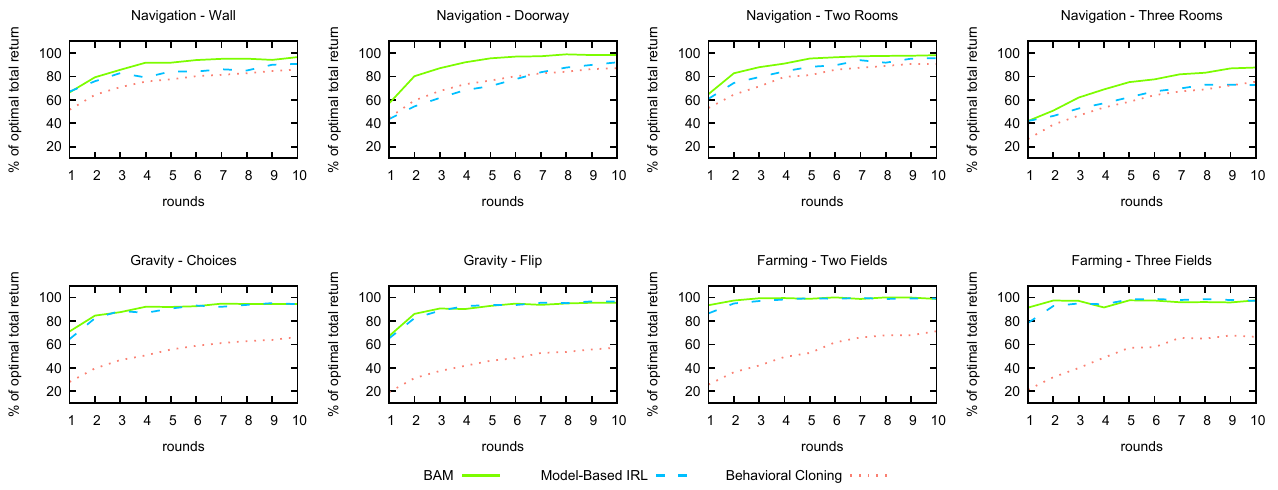}
    	\caption{{\small The total return (averaged over 50 episodes) of the policies learned by BAM, model-based IRL, and behavioral cloning, as a percentage of the total return for the optimal policies, learning from demonstrations and feedback combined.  Curves are averages over 50 separate agents learning from scratch.}}
    	\label{fig:performance_combined}
    \end{figure*}
    
    One of the main advantages of the BAM algorithm in terms of reducing teacher effort is that it is able to share a teacher's knowledge across multiple tasks through the dynamics model.  An alternative (and potentially simpler) approach to sharing such information would be to learn a global cost function in addition to the task specific cost functions, such that the policies learned for each task would be optimal for the sum of the global and task costs.  The global cost function might encode much of the same teacher knowledge that BAM captures with its dynamics model.  For example, in the navigation domain, unobserved obstacles could be represented as states with a high global cost, with goals captured as low-cost states in the task-specific cost functions.
    
    To understand the benefits of learning a global dynamics model, we compared a version of model-based IRL that learned a global cost function against both BAM and model-based IRL without global costs.  Figure~\ref{fig:performance_common} shows the results of these comparisons, using the same training protocol and environments as the experiments in Section~\ref{sec:sim_results}.  We can see that in most cases, the use of a global cost function offers little advantage over the model-based IRL algorithm, that is, the total return for the global cost algorithm is less than or equal to that of model-based IRL.  In fact, we find that using a global cost function actually hurts performance in the gravity domain.  This reflects the fact that a global cost function cannot capture the outcomes of unseen actions, only the immediate costs.  The fact that BAM dominates the global cost algorithm in the gravity and farming domains can be explained by noting that the space of global cost functions is much less constrained (and has many more parameters) than the space of dynamics models, such that the agent requires less data to learn a good dynamics model than it does a good global cost function.  The simpler space of dynamics models is possible because we have domain-specific prior knowledge of the possible dynamics, knowledge which would be difficult to incorporate into a space of global cost functions.
    
    \subsection{Demonstrations and Feedback}
    
    As we are interested in the case where the teacher uses a combination of demonstrations and evaluative feedback, we also conducted a set of experiments in which, for each round, the agent first observed a set of demonstrations of each task, and then attempted each task on its own while the simulated teacher provided feedback.  These experiments (Figure~\ref{fig:performance_combined}) show a major improvement for model-based IRL in most domains, though BAM still has an advantage in the navigation environments. The improvements for model-based IRL the farming and gravity domains largely reflect the fact that the agent can now explore the environment under its own policies, and can therefore directly observe transitions that would not have been seen during the teacher's optimal demonstrations.  This exploration is less valuable in the navigation domain, however, as the space of dynamics models is much more complex.  Inferring the dynamics from the teacher's behavior is still more efficient in this domain than requiring that each obstacle be directly observed.  As learning from feedback only (no demonstrations) in our domains proved to be very inefficient for all three algorithms, we did not evaluate BAM when learning from feedback alone.

\section{Human-Subjects Experiments}
\label{sec:humans}

    \begin{table*}[t]
        \centering
        \caption{Results of the Mechanical Turk user study.  Learning sessions were evaluated on whether the agent's policies reached 50\% and 80\% of the optimal total return, and on the number of episodes and the number of actions (under either the teacher or the agent's control) required on average to reach these thresholds.  (*) indicates that a value is significantly worse ($p \leq 0.05$, t-test or Fisher's exact test) than the value for BAM, while ($\dagger$) indicates that the result remains significant under a Benjamini-Yekutieli~\cite{benjamini2001fdr} correction for multiple comparisons (with a false discovery rate of 20\%). \vspace{0.2cm}}
        \begin{tabular}{l|r|r|r|r|r|r|r|}
            & {\bf Total Sessions} & \multicolumn{2}{c|}{\bf Successful} & \multicolumn{2}{c|}{\bf Episodes Required} & \multicolumn{2}{c|}{\bf Actions Required} \\ \hline
            {\bf Two Rooms}    &    & {\bf 50\%}   & {\bf 80\%}  & {\bf 50\%}     & {\bf 80\%}    & {\bf 50\%}      & {\bf 80\%}     \\ \hline
            BAM                & 27 & 24           & 14          & 7.0            & 9.9           & 164.2           &  215.9         \\
            Model-Based IRL    & 27 & 25           & *6          & 7.4            & 12.8          & 168.1           & 258.5          \\
            Behavioral Cloning & 27 & 17           & *$\dagger$1 & 8.4            & 20.0          & 207.3           & 461.0          \\ \hline
            {\bf Doorway}      &    & {\bf 50\%}   & {\bf 80\%}  & {\bf 50\%}     & {\bf 80\%}    & {\bf 50\%}      & {\bf 80\%}     \\ \hline
            BAM                & 27 & 19           & 5           & 9.7            & 14.7          & 167.4           & 217.6          \\
            Model-Based IRL    & 27 & 18           & 4           & *12.9          & *26.6         & 215.4           & *441.2         \\
            Behavioral Cloning & 27 & 12           & 4           & *$\dagger$13.7 & *27.7         & 271.0           & *517.7         \\ \hline
            {\bf Two Fields}   &    & {\bf 50\%}   & {\bf 80\%}  & {\bf 50\%}     & {\bf 80\%}    & {\bf 50\%}      & {\bf 80\%}     \\ \hline
            BAM                & 31 & 27           & 18          & 2.8            & 3.6           & 61.4            & 87.6           \\
            Model-Based IRL    & 31 & 27           & 12          & *$\dagger$4.4  & *$\dagger$6.5 & *$\dagger$103.1 & 131.0          \\
            Behavioral Cloning & 31 & *$\dagger$10 & *$\dagger$0 & *$\dagger$8.4  & N/A           & *272.1          & N/A            \\ \hline
            {\bf Three Fields} &    & {\bf 50\%}   & {\bf 80\%}  & {\bf 50\%}     & {\bf 80\%}    & {\bf 50\%}      & {\bf 80\%}     \\ \hline
            BAM                & 31 & 24           & 20          & 1.3            & 1.6           & 26.9            & 29.8           \\
            Model-Based IRL    & 31 & 22           & 12          & *1.8           & *$\dagger$2.4 & *$\dagger$38.3  & *$\dagger$53.0 \\
            Behavioral Cloning & 31 & 17           & *$\dagger$2 & *$\dagger$2.7  & 7.5           & *$\dagger$62.8  & *201.5         \\ \hline
        \end{tabular}
        \label{tab:mturk_study}
    \end{table*}

    As the purpose of this work is to allow non-expert humans to teach agents with less effort than is possible with existing approaches, we need to determine whether the BAM algorithm actually reduce this effort for real human teachers.  We therefore compared BAM against the model-based IRL and behavioral cloning algorithms from Section~\ref{sec:alternatives} in a web-based user study in which participants each trained several learning agents in the four environments shown in Figure~\ref{fig:environments_human}.  Participants were recruited through the Amazon Mechanical Turk platform, and were paid \$3.00 US for completing the experiment.  Each participant was randomly assigned to either the navigation or farming domain, and went through a brief tutorial showing them how to use the training interface in that domain.  Participants assigned to the navigation domain first taught three agents in the Two Rooms environment, and then taught three agents in the Doorway environment.  Those assigned to the farming domain first taught agents in the Two Fields environment, and then in the Three Fields environment.  Between both domains we had a total of 58 unique participants who completed the study, out of 129 (possibly non-unique) participants who began the experiment.  These participants spent an average of 33 minutes (std. deviation 17 minutes) working on the study (including the tutorial).

    \subsection{Learning Sessions}
    
    Participants taught one agent at a time, and could not return to an agent after moving on to the next one.  The interface shown to the participants allowed them to take control of the agent and demonstrate a task, or ask the agent to try and perform the task itself.  During demonstrations, the participant controlled the agent using the arrow keys to move up, down, left, or right.  While the agent was acting on its own, participants had the option of providing positive and negative feedback using the keyboard as well.  Participants were prompted to reset the environment to a random initial state after a task was completed, and had the option of moving the agent to desired start locations themselves.  The participant could switch between tasks at any time, and the agent would associate any demonstrated action or feedback signal with the currently selected task.  Participants were required to provide at least one demonstrated action for each task before being allowed to move on to teaching the next agent.

    We compared the BAM algorithm against both behavioral cloning and model-based IRL.  Participants were asked to train three agents in both of the environments they encountered, with each agent being controlled by a different learning algorithm.  The order in which these algorithms were presented was randomized for each participant, and each agent was rendered with a different color to highlight the fact that it did not know what the previous agents had learned. The learning agents themselves were configured identically to the ones used in the simulation experiments (see Section~\ref{sec:simulation}).  Learning updates only occurred after the participant ended a demonstration or an evaluation episode.  Because the teacher had no way to demonstrate remaining in the same location, at the end of a demonstration a synthetic no-op action was shown to the agent to allow it to identify goal states.
        
    \subsection{Results}
    
    Table~\ref{tab:mturk_study} shows the number of learning sessions in which the total return of the agent's policies reached 50\% and 80\% of the optimal total return, as well as the average number of episodes (including demonstrations and agent-controlled episodes) and individual actions (both teacher and agent actions) required for the agent to reach these thresholds.  The total return was estimated by running each of the agent's learned policies for 1000 simulated episodes.  We can see in Table~\ref{tab:mturk_study} that BAM dominates the alternative algorithms in almost every case, save for the 50\% threshold in the Two Rooms environment.  For the 80\% threshold in particular, BAM reduces the number of actions and episodes required across all four environments, and is more likely to reach 80\% of the optimum.  Multi-way ANOVA's (with algorithm, environment and session order as factors) show that BAM's advantages over model-based IRL and behavioral cloning in terms of the number of episodes and actions needed to reach the 50\% and 80\% thresholds are significant ($p < 0.01$).  Fisher's exact test's (for all environments combined) show that BAM's superior success rates versus the alternatives at the 80$\%$ threshold are also significant ($p < 0.01$).  
    
    More specifically, in the Two Rooms environment (rows 3 to 5), Fisher's exact test shows that the difference in the number of sessions using BAM (14 sessions) and model-based IRL (6 sessions) that reached the 80\% threshold is significant ($p = 0.047$).  In the Doorway environment (rows 7 to 9), t-tests also show that BAM requires significantly ($p < 0.05$) fewer actions and episodes than the alternatives to reach the $80\%$ threshold.  To address multiplicity, we also perform a Benjamini-Yekutieli~\cite{benjamini2001fdr} correction on all of the environment-specific comparisons.  Though some results are no longer significant under this correction, BAM's advantages over model-based IRL in terms of the number of episodes required at the $80\%$ threshold (column 6), and in terms of the number of actions required to reach the 50\% threshold (column 7), are still significant in both the Two Fields and Three Fields environments.  These results demonstrate that BAM is more efficient and more reliable in learning from human teachers than model-based IRL or behavioral cloning, meaning that an agent using BAM can learn more complex sets of behaviors than the alternatives, with no additional effort on the part of the teacher.

\section{Conclusions}
\label{sec:conclusions}

    In this work, we have presented a novel approach to interactive learning that takes full advantage of a teacher's understanding of the learning agent's environment.  We have shown that BAM dominates existing approaches (which ignore the teacher's dynamics knowledge) across many different domains and measures of teacher effort.  Future work will focus on scaling BAM to more complex problems.  This includes replacing exact value iteration with sparse search in BAM's planning model, along the lines of~\cite{macglashan2015receding}.  In many domains however, building a sufficiently accurate one-step model may be difficult, and so we are also interested in finding compact, high-level dynamics models (e.g. the value iteration networks developed in~\cite{tamar2016vin}) that capture the teacher's knowledge of the environment.   Finally, we are particularly interested in implementing BAM on physical robots, a key application area for interactive learning.  Overall, we are extremely encouraged by BAM's performance in reducing teacher workload both in simulation and when learning from real humans, and our results provide a strong foundation for future work applying this approach to real-world domains.

\bibliographystyle{aaai}
\bibliography{references}

\onecolumn
\include{appendices}

\end{document}

%% file: appendices.tex
\begin{appendices}

\section{User Interface for Human-Subjects Experiments}
\label{apx:interface}

    \begin{figure}[!h]
        \centering
        \includegraphics[width=0.9\columnwidth]{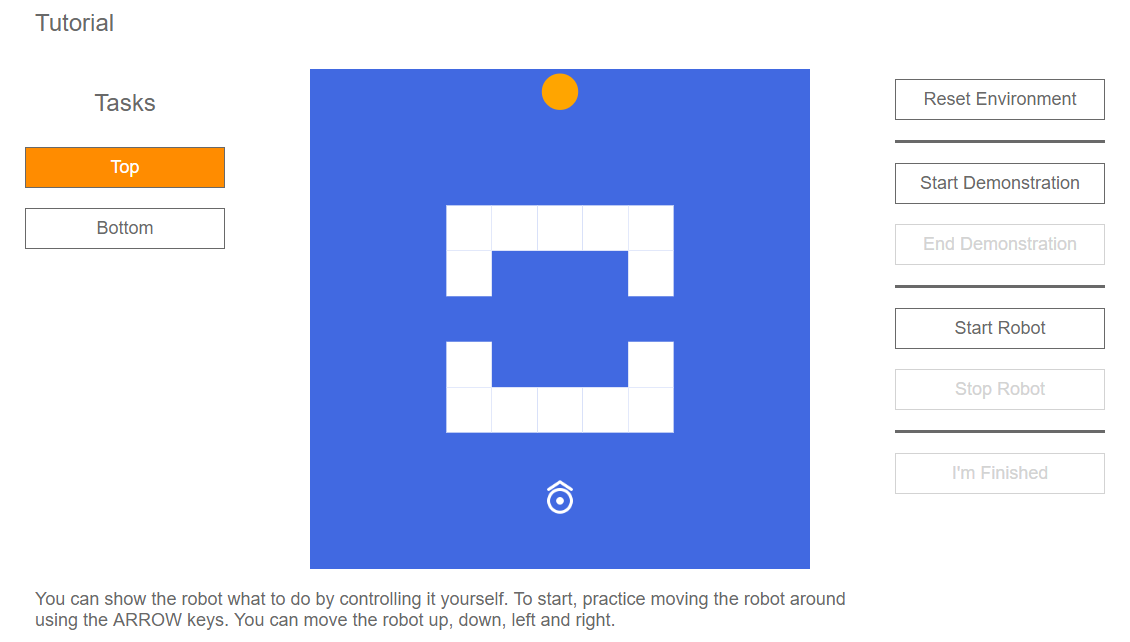}
        \captionof{figure}{{\small A screen shot of the user interface for the user study conducted through Amazon Mechanical Turk.  The interface is currently in the tutorial mode for the navigation domain.}}
        \label{fig:screenshot_navigation}
    \end{figure}

    \begin{figure}[!h]
        \centering
        \includegraphics[width=0.9\columnwidth]{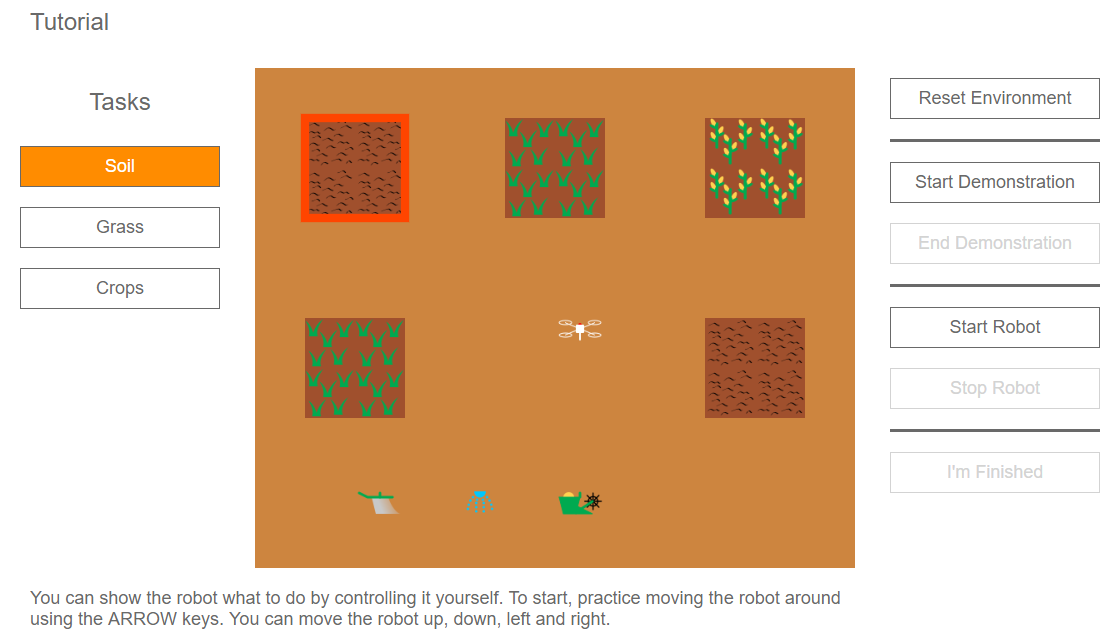}
        \captionof{figure}{{\small A screen shot of the user interface for the user study conducted through Amazon Mechanical Turk.  The interface is currently in the tutorial mode for the farming domain.}}
        \label{fig:screenshot_farming}
    \end{figure}

    \pagebreak[4]

\section{Experimental Environments}
\label{apx:environments}

    \begin{figure}[!h]
        \centering
    	\subfloat[{\footnotesize Navigation - Wall}]{
    		\includegraphics[width=0.45\columnwidth]{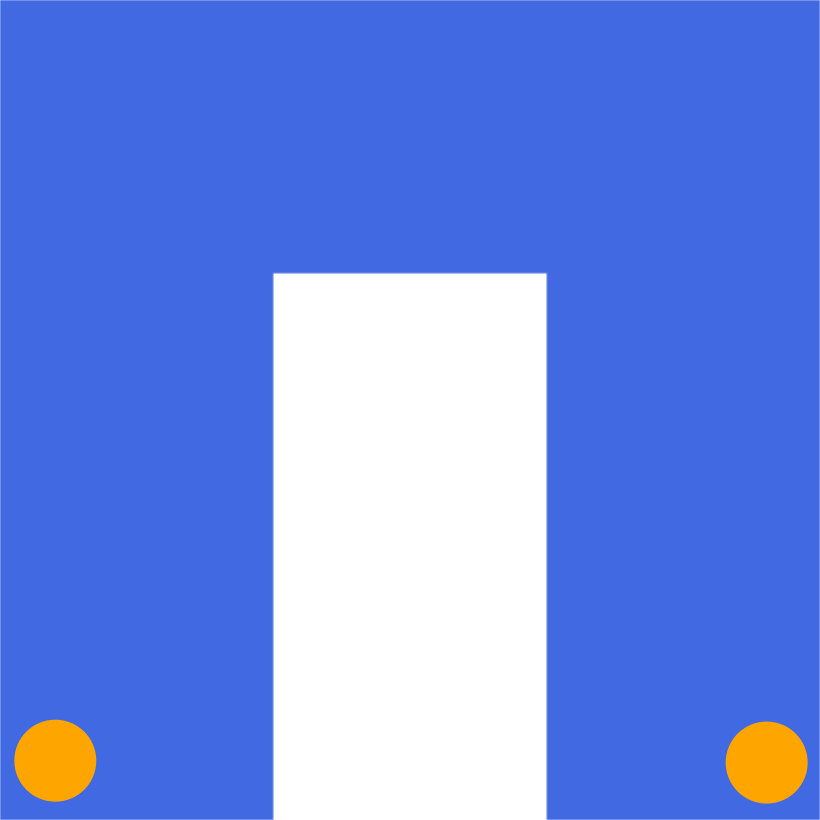}
    	}
    	\subfloat[{\footnotesize Navigation - Doorway}]{
    		\includegraphics[width=0.45\columnwidth]{figures/environments/doors_goals.png}
    	} \break
    	\subfloat[{\footnotesize Navigation - Two Rooms}]{
    		\includegraphics[width=0.45\columnwidth]{figures/environments/two_rooms_goals.png}
    	}
    	\subfloat[{\footnotesize Navigation - Three Rooms}]{
    		\includegraphics[width=0.45\columnwidth]{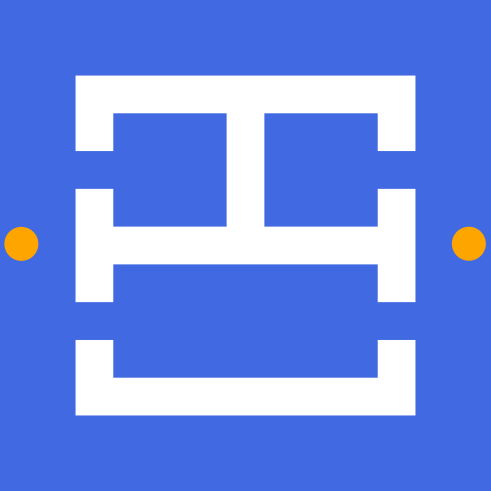}
    	}
    	\caption{{\small The four navigation environments used in the simulated teacher experiments, including the Doorway and Two Rooms environments used in the human subjects experiments.  Orange circles indicate goal locations, with each goal defining a different task.  White squares indicate states blocked by obstacles.}}
    	\label{fig:navigation_environments}
    \end{figure}

    \pagebreak[4]
    
    \begin{figure}[!h]
        \centering
    	\subfloat[{\footnotesize Gravity - Flip}]{
    		\includegraphics[width=0.35\columnwidth]{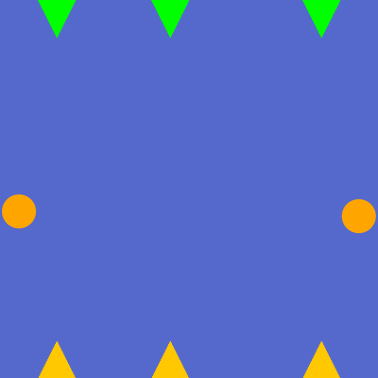}
    	}
    	\subfloat[{\footnotesize Gravity - Choices}]{
    		\includegraphics[width=0.45\columnwidth]{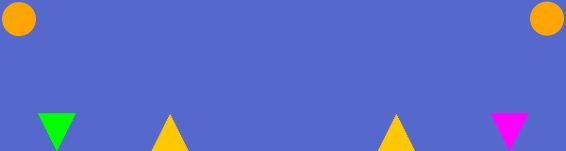}
    	}
    	\caption{{\small The two gravity environments used in the simulated teacher experiments.  Orange circles indicate goal locations, with each goal defining a different task.  Arrows indicate states that change the direction of the gravity, but the agent can only see the color of these arrows, not their direction.  The unknown dynamics consist of the mapping from colors to gravity directions.}}
    	\label{fig:gravity_environments}
    \end{figure}

    \begin{figure}[!h]
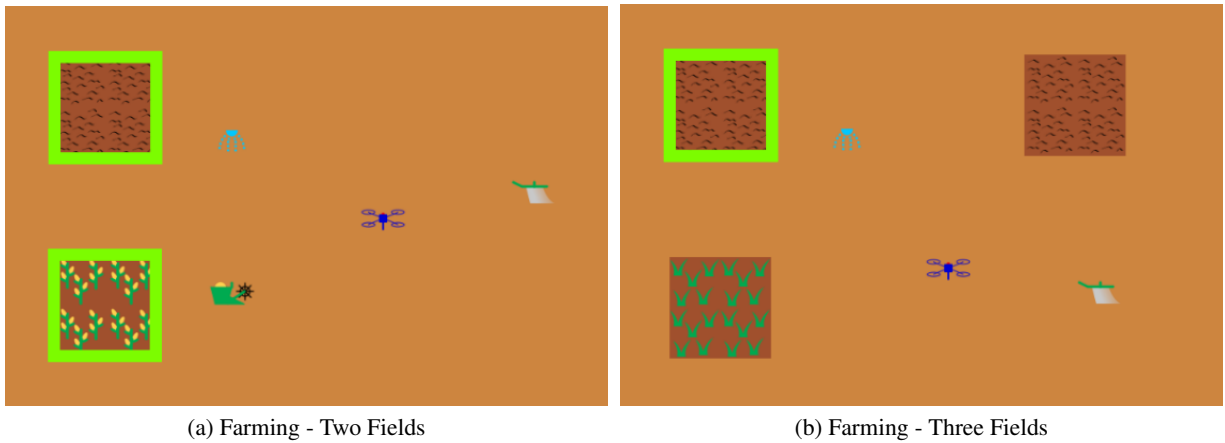

        \centering
    	\subfloat[{\footnotesize Farming - Two Fields}]{
    		\includegraphics[width=0.45\columnwidth]{figures/environments/two_fields_goals.png}
    	}
    	\subfloat[{\footnotesize Farming - Three Fields}]{
    		\includegraphics[width=0.45\columnwidth]{figures/environments/three_fields_goals.png}
    	}
    	\caption{{\small The environments used in both the simulated teacher and human subjects experiments.  Target Fields are highlighted with green squares, with each target field defining a different task.  Also visible are the agent itself (the blue drone), and the three farm implements (only the plow and sprinkler are available in (b)).}}
    	\label{fig:farming_environments}
    \end{figure}

    \pagebreak[4]

\section{Simulation Results}

    \begin{center}
        \includegraphics[width=\columnwidth]{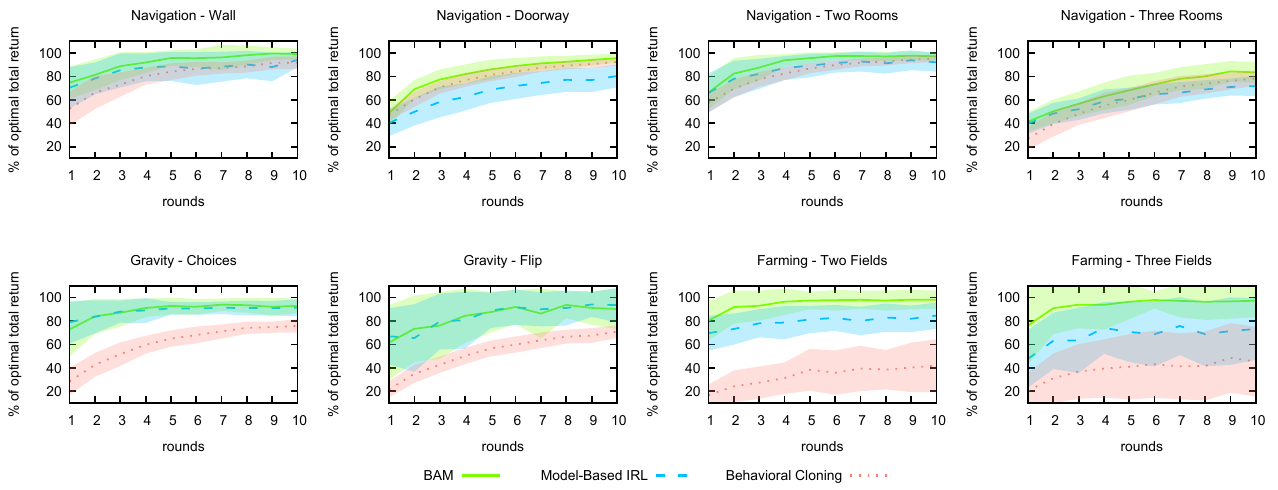}
        \captionof{figure}{{\small The total return (averaged over 50 episodes) of the policies learned by BAM, model-based IRL, and behavioral cloning, as a percentage of the total return for the optimal policies.  Total return is the sum of the returns for each task.  Curves are averages over 50 separate agents learning from scratch.  Shaded regions show the standard errors of the means.}}
        \label{fig:performance_error}
    \end{center}

    \begin{center}
        \includegraphics[width=\columnwidth]{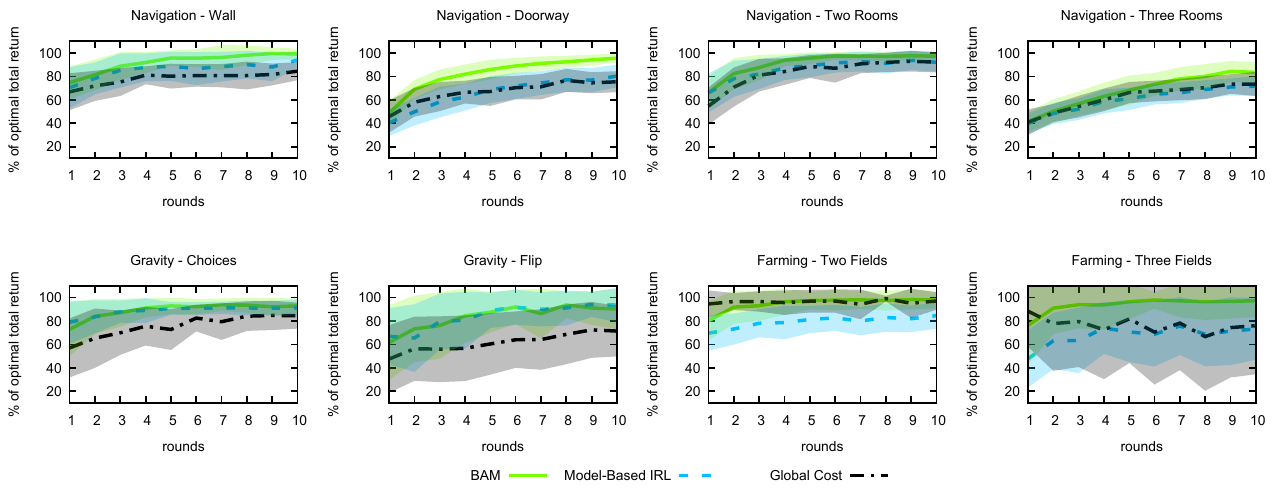}
        \captionof{figure}{{\small The total return (averaged over 50 episodes) of the policies learned by BAM, model-based IRL, and model-based IRL with global costs, as a percentage of the total return for the optimal policies.  Total return is the sum of the returns for each task.  Curves are averages over 50 separate agents learning from scratch.   Shaded regions show the standard errors of the means.}}
        \label{fig:performance_common_errors}
    \end{center}
    
    \begin{center}
        \includegraphics[width=\columnwidth]{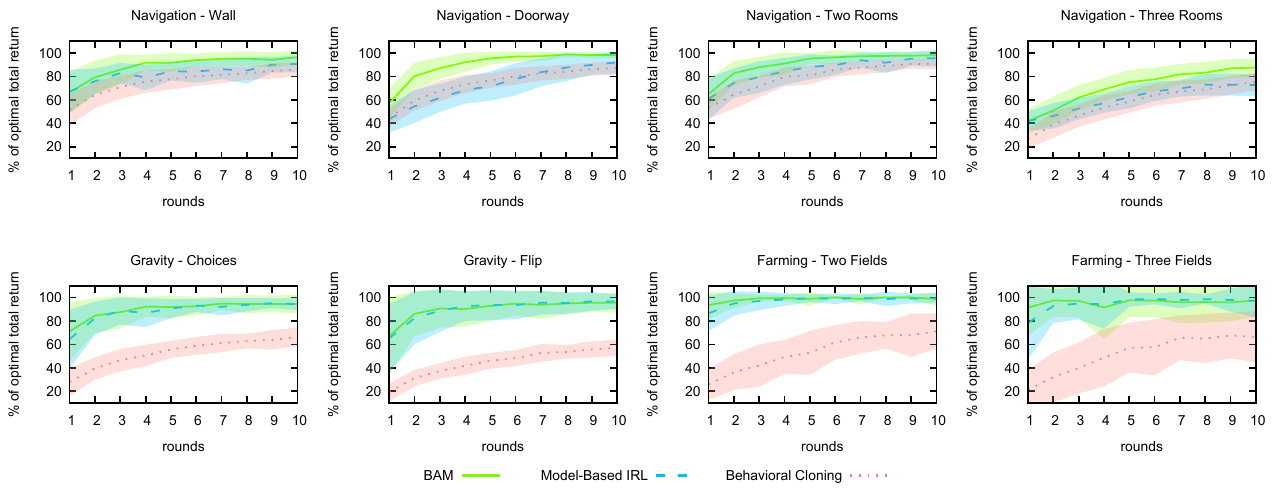}
        \captionof{figure}{{\small The total return (averaged over 50 episodes) of the policies learned by BAM, model-based IRL, and behavioral cloning, as a percentage of the total return for the optimal policies, learning from demonstrations and feedback combined.  Curves are averages over 50 separate agents learning from scratch.   Shaded regions show the standard errors of the means.}}
        \label{fig:performance_combined_errors}
    \end{center}
    
    \pagebreak[4]

\section{Statistical Analysis of Human subjects Experiments}

    \begin{table*}[!h]
        \centering
        \caption{P-values for comparisons done across all environments.  For each of the six measures of performance, we compare BAM against model-based IRL, and against behavioral cloning. For the numbers of episodes and actions, we use multi-way ANOVA's with factors for the algorithm and the environment used for a session, and for the position of that session in the sequence of sessions the participant completed.  For the success rates, we use Fisher's exact test, and treat the environment as a latent value, but one which we know is independent of the choice of algorithm.}
        \begin{tabular}{l|l|r|r|}
        {\bf Performance Measure} & {\bf Threshold} & {\bf Model-Based IRL} & {\bf Behavioral Cloning} \\ \hline
        \multirow{2}{*}{\bf Success Rate}      & {\bf 50\%} & 0.4276   & 3.926e-10 \\
                                               & {\bf 80\%} & 1.298e-4 & 2.2e-16 \\ \hline
        \multirow{2}{*}{\bf Episodes Required} & {\bf 50\%} & 6.47e-4  & 1.25e-12 \\
                                               & {\bf 80\%} & 9.38e-06 & $<$ 2e-16 \\ \hline
        \multirow{2}{*}{\bf Actions Required}  & {\bf 50\%} & 0.00376  & 1.20e-10 \\
                                               & {\bf 80\%} & 1.92e-4  & 1.34e-14  \\ \hline
        \end{tabular}
        \label{tab:analysis_all}
    \end{table*}
    
    \begin{table*}[!h]
        \centering
        \caption{Raw p-values for the environment specific comparisons against BAM.  For the success rates in each environment, Fisher's exact test is used, while unpaired t-tests are used for the number of episodes and actions required.}
        {\footnotesize
        \begin{tabular}{l|l|l|r|r|}
        {\bf Environment} & {\bf Performance Measure} & {\bf Threshold} & {\bf Model-Based IRL} & {\bf Behavioral Cloning} \\ \hline 
        \multirow{6}{*}{{\bf Two Rooms}} & \multirow{2}{*}{\bf Success Rate}      & {\bf 50\%} & 1            & 5.367713e-02 \\
                                         &                                        & {\bf 80\%} & 4.727039e-02 & 1.291742e-04 \\
                                         & \multirow{2}{*}{\bf Episodes Required} & {\bf 50\%} & 0.7058005    &  0.2355695 \\
                                         &                                        & {\bf 80\%} & 0.11362855   & N/A \\
                                         & \multirow{2}{*}{\bf Actions Required}  & {\bf 50\%} & 0.9153665    & 0.1368809 \\
                                         &                                        & {\bf 80\%} & 0.3346513    & N/A \\ \hline
        \multirow{6}{*}{{\bf Doorway}}   & \multirow{2}{*}{\bf Success Rate}      & {\bf 50\%} & 1            & 9.777568e-02 \\
                                         &                                        & {\bf 80\%} & 1            & 1 \\
                                         & \multirow{2}{*}{\bf Episodes Required} & {\bf 50\%} & 4.159614e-02 & 7.416946e-03 \\
                                         &                                        & {\bf 80\%} & 3.728851e-02 & 4.383471e-02 \\
                                         & \multirow{2}{*}{\bf Actions Required}  & {\bf 50\%} & 0.1120207    & 5.307102e-02 \\
                                         &                                        & {\bf 80\%} & 4.483479e-02 & 4.535574e-02 \\ \hline
        \multirow{6}{*}{{\bf Two Fields}} & \multirow{2}{*}{\bf Success Rate}      & {\bf 50\%} & 1            & 4.309968e-06 \\
                                          &                                        & {\bf 80\%} & 0.2035422    & 2.230871e-07 \\
                                          & \multirow{2}{*}{\bf Episodes Required} & {\bf 50\%} & 1.252016e-04 & 9.592095e-05 \\
                                          &                                        & {\bf 80\%} & 1.197617e-04 & N/A \\
                                          & \multirow{2}{*}{\bf Actions Required}  & {\bf 50\%} & 4.849809e-03 & 1.767247e-03 \\
                                          &                                        & {\bf 80\%} & 7.011737e-02 & N/A \\ \hline
        \multirow{6}{*}{{\bf Three Fields}} & \multirow{2}{*}{\bf Success Rate}      & {\bf 50\%} & 1            & 6.200178e-02 \\
                                            &                                        & {\bf 80\%} & 4.130673e-02 & 2.376083e-06 \\
                                            & \multirow{2}{*}{\bf Episodes Required} & {\bf 50\%} & 1.985265e-02 & 5.721056e-04 \\
                                            &                                        & {\bf 80\%} & 1.475193e-03 & 0.1536753 \\
                                            & \multirow{2}{*}{\bf Actions Required}  & {\bf 50\%} & 4.085855e-02 & 2.319712e-03 \\
                                            &                                        & {\bf 80\%} & 4.712152e-03 & 2.819027e-02 \\ \hline
        \end{tabular}}
        \label{tab:analysis_raw}
    \end{table*}

\end{appendices}